\documentclass[runningheads]{llncs}
\usepackage{graphicx}
\usepackage{amsmath}
\usepackage{amssymb}
\usepackage{subfig}
\usepackage{cite}
\usepackage{booktabs}
\usepackage{xcolor}
\usepackage{multirow}
\usepackage{todonotes}
\usepackage{tabularx}
\usepackage{url}
\usepackage{comment}
\usepackage{adjustbox}
\usepackage[ruled,vlined]{algorithm2e}
\usepackage{enumerate}
\usepackage{appendix}

\newcommand{\printfnsymbol}[1][\value{footnote}]{\footnotemark[#1]}
\usepackage{authblk}
\begin{document}
\title{\textbf{GLOWin: A  Flow-based Invertible Generative Framework for Learning Disentangled Feature Representations in Medical Images}}

\titlerunning{GLOWin: A Flow-based Model for Disentangled Representations.}

\author{Aadhithya Sankar\inst{1,}\thanks{\textit{First two authors contributed equally to this work.
\newline $^\dagger$ S. T. Kim and N. Navab share senior authorship.
\newline $^{\ast}$ Corresponding author (st.kim@khu.ac.kr)
}}, Matthias Keicher\inst{1,}\printfnsymbol, Rami Eisawy\inst{2,3}, Abhijeet Parida\inst{2}, Franz Pfister\inst{2,4}, Seong Tae Kim\inst{5,\dagger,\ast}, Nassir Navab\inst{1,6,\dagger}}
\authorrunning{A. Sankar and M. Keicher et al.}

\institute{Computer Aided Medical Procedures, Technical University of Munich, Germany
\and
deepc GmbH
\and
Technical University of Munich, Germany
\and
Ludwig Maximilians University Munich, Germany
\and
Department of Computer Science and Engineering, Kyung Hee University, Korea
\and
Computer Aided Medical Procedures, Johns Hopkins University, USA
}

\maketitle              

\begin{abstract}
Disentangled representations can be useful in many downstream tasks, help to make deep learning models more interpretable, and allow for control over features of synthetically generated images that can be useful in training other models that require a large number of labelled or unlabelled data. Recently, flow-based generative models have been proposed to generate realistic images by directly modeling the data distribution with invertible functions. In this work, we propose a new flow-based generative model framework, named GLOWin, that is end-to-end invertible and able to learn disentangled representations. Feature disentanglement is achieved by factorizing the latent space into components such that each component learns the representation for one generative factor. Comprehensive experiments have been conducted to evaluate the proposed method on a public brain tumor MR dataset. Quantitative and qualitative results suggest that the proposed method is effective in disentangling the features from complex medical images.

\end{abstract}
\section{Introduction}
\label{sec:intro}
A disentangled representation can be defined as one where single latent units are sensitive to changes in single generative factors, while being relatively invariant to changes in other factors \cite{betavae}. Disentangled representations are more interpretable and robust\cite{bengio2013representation} and can help to solve some challenges in deep learning for medical imaging such as the lack of adequate labelled data and class imbalance by generating new samples with specific features by manipulating the representations of existing samples in addition to increased interpretability of the models\cite{Litjens_2017}.
Various Variational Autoencoder\cite{kingma2013autoencoding} based methods \cite{betavae, kim2018disentangling, chen2018isolating} have been proposed, which perform disentanglement by introducing constraints to the Kullback-Leibler divergence regularization term.   Esser et al. introduce Interpretable Invertible Networks (IIN) \cite{esser2020disentangling} where they use a flow-based invertible network as an add-on model to learn to disentangle the latent spaces of pre-trained autoencoders or classifier models. Flow-based generative models \cite{dinh2014nice,dinh2016density,kingma2018glow}, unlike Variational Autoencoders(VAEs)\cite{kingma2013autoencoding} and Generative Adversarial Networks(GANs)\cite{goodfellow2014generative} can directly model the data distribution, compute exact log-likelihood with a series of invertible and bijective transformations and have also shown to learn good representations\cite{jacobsen2018revnet, kingma2018glow}. 

In this work, we adapt the Factor  Loss  proposed  in  [8]  to work with spatial embeddings and  the  Glow  model  [15]  and  propose  a  new flow-based generative model framework, named GLOWin, that is able to learn disentangled representations by factorizing the latent space in to factors. Unlike the IIN\cite{esser2020disentangling} model which is trained on the embeddings from a pre-trained network, our model is directly trained on the images, thereby getting rid of the need for pre-trained embeddings. More specifically, we train the model on the Multimodal Brain Tumor Segmentation Challenge (BraTS 2019) dataset 
\cite{menze2014multimodal, bakas2017advancing,bakas2019identifying} and factorize the latent space into three factors: anomaly, slice index and the residual. The anomaly factor learns the representation for the presence of anomaly in the input slice. The slice index denotes the location of the slice in the axial plane. Since the skull-stripped brain images are co-registered to the same anatomical template, slice index acts as a proxy label to the approximate shape of the brain in each slice. Therefore, the slice index factor might be able to learn the representation for the anatomical location and shape of the brain in particular slices. Finally, the residual learns the representation of all other semantic concepts. 
Our approach is different from existing approaches in the medical imaging domain in a few ways. Existing approaches in the medical domain\cite{Chartsias_2019, jiang2020semisupervised, qin2019unsupervised}  learn separate representations for each factor using separate encoders and loss functions. Our model on the contrary adopts the Factor Loss\cite{esser2020disentangling} and learns disentangled representations by factorizing the latent embedding into factors.

In short, the key contributions of the proposed flow-based generative model framework are as follows:
\begin{itemize}
    \item It is able to learn disentangled representations by factorizing the latent space into meaningful factors, obtaining a disentangled latent space without needing separate losses for each factor.
    \item It is able to generate images with or without predetermined characteristics enabling for example the introduction of a particular class of anomalies into healthy slices using the proposed disentangled representations.
    \item It is end-to-end invertible and uses a novel adaptation of the Factor Loss to work with spatial embeddings.
\end{itemize}
We also conduct comprehensive evaluation on the disentangled factors in the downstream tasks and analysis on the robustness of the model. 

\section{Methodology}\label{sec:methodology}
\subsubsection{Glow with interpretable latent space (GLOWin).}  Our model extends the Glow\cite{kingma2018glow} model by adopting the Factor Loss \cite{esser2020disentangling} to learn disentangled factors.
For a flow-based model, the log probability density of the data distribution is computed using the \textit{change of variables theorem} as

\begin{equation}
    \log p(\vec{x}) = \log\, p(\vec{z})+\sum_{i=1}^{K} \log\left|\det\, \frac{d\vec{h}_i}{d\vec{h}_{i-1}} \right|\label{eq:flow_loglikelihood}
\end{equation}
where $\vec{h}_i = f_i(\vec{h}_{i-1}), \vec{h}_0 = \vec{x}, \vec{h}_K = \vec{z}$.
The scalar value $\log\left|\det\, \frac{d\vec{h}_i}{d\vec{h}_{i-1}} \right|$ is the log determinant.

We use a multi-scale architecture as specified in \cite{dinh2016density}, affine coupling and invertible $1\times1$ convolutions for the channel permutation. LU decomposition are used to initialize the weights of the $1\times1$ convolutions for efficient computation. We have a total of $L$ scales and each scale has $K$ flow steps.
As discussed in \cite{dinh2016density}, to keep the computational and memory cost low, at each scale half the dimensions of the output is factored out, i.e.,
\begin{align}
    (\vec{z}_{i+1}, \vec{h}_{i+1}) = f_{i+1}(h_i) \label{eq:multi_scale_rep}\\
    \vec{z}_L = f_L(h_{L-1}) \label{eq:multi_scale_last}\\
    \vec{z} = (z_1, z_2, ..., z_L) \label{eq:multi_scale_zx}
\end{align}

\subsubsection{Feature disentanglement loss.}\label{sec:mtd_glowin_loss} We model the factored out embeddings from each scale as a unit gaussian, i.e.,
\begin{equation}\label{eq:z1l_unit}
\mathbf{z}_{0\dotsb L-1} \sim \mathcal{N}(\mathbf{0},\mathbf{I})
\end{equation}

Disentanglement of features is enforced in the last embedding output $\mathbf{z}_L \in \mathbb{R}^{C\times H\times W}$ and to achieve this, we adopt the method proposed in \cite{esser2020disentangling} for our model. The embedding $\mathbf{z}_L$ is factorized into components along the channel dimension such that
\begin{equation}
    \label{eq:zl_fac}
    \mathbf{\tilde{z}}_L = (\mathbf{z}_L^{m})_{m=0}^{M}
\end{equation}
where $\mathbf{z}_L^m \in \mathbb{R}^{C_m \times H \times W}\nonumber$ and $\sum_{m=0}^{M}C_m = C$. $\mathbf{z}_L$ is factorized to $M+1$ factors, where the $M$ factors correspond to the $M$ semantic features we want to disentangle and the remaining $1$ factor acts as the residue, learning features that do not correspond to any of semantic features we are disentangling.
To modify the factors independent of each other, we model the factorized embedding to be a joint distribution of the individual factors, where each factor themselves are modelled as unit normal distribution, i.e.,
\begin{equation}
    \label{eq:parts_eq}
    p(\mathbf{\tilde{z}}_L) = \prod_{m=0}^{M} \mathcal{N}(\mathbf{z}_{L}^{m};\mathbf{0},\mathbf{I})
\end{equation}

We apapt the Factor Loss from \cite{esser2020disentangling} to work with spatial feature maps instead of non-spatial vector embeddings. For any given training image pair $(\mathbf{x}_a, \mathbf{x}_b) \sim p(\mathbf{x}_a,\mathbf{x}_b;F)$ sharing the same semantic concept $F$, we want to ensure that the corresponding factor $\mathbf{z}_L^F$ for the pair lie close to each other and the images are invariant with the other factors. This is achieved by using a positive correlation factor $\sigma_{ab} \in (0,1)$ for the $F$th factor for the pair as
\begin{equation}
    \mathbf{z}_{L_b}^F \sim \mathcal{N}(\sigma_{ab}\mathbf{z}_{L_a}^F, (1-\sigma_{ab}^2)\mathbf{I}) \label{eq:pos_cor}
\end{equation}
and for the other uncorrelated factors,
\begin{equation}
    \label{eq:un_cor}
    \mathbf{z}_{L_b}^m \sim \mathcal{N}(\mathbf{0},\mathbf{I}), \: m \in \{0,1,...,M\} \setminus \{F\}
\end{equation}

\begin{figure}[t]
\centering
\includegraphics[width=0.75\textwidth]{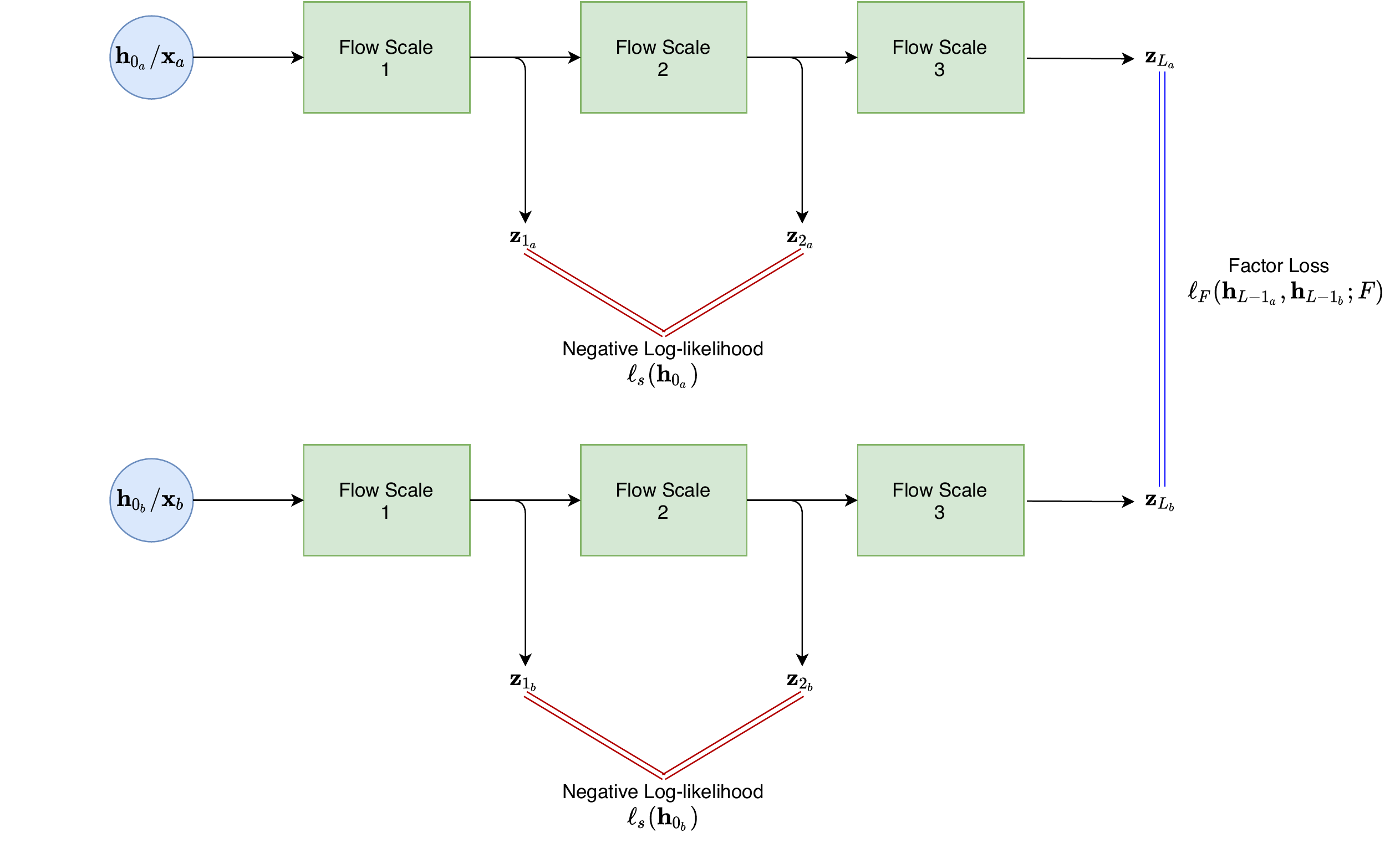}
\caption{Our model is trained using image pairs sharing the semantic concept to be disentangled. The Factor Loss is applied only to the last latent to achieve disentanglement.} \label{fig:glowin_loss}
\end{figure}

From \eqref{eq:flow_loglikelihood}, the likelihood for the image pair $(\vec{x}_a, \vec{x}_b)$ for the last scale $L$ can be calculated as,
\begin{align}
p(\mathbf{h}_{L-1_a},&\mathbf{h}_{L-1_a};F) = p(\mathbf{h}_{L-1_a})\cdot p(\mathbf{h}_{L-1_b};\mathbf{h}_{L-1_b},F)\nonumber\\
&=p(f_L(\mathbf{h}_{L-1_a}))\cdot \left|\det f'_{L}(\mathbf{h}_{L-1_a})\right|\nonumber\\
&\cdot p(f_L(\mathbf{h}_{L-1_b});f_L(\mathbf{h}_{L-1_a},F))
\cdot \left|\det f'_{L}(\mathbf{h}_{L-1_b})\right|\label{ll5}
\end{align}
Substituting \eqref{eq:parts_eq}, \eqref{eq:pos_cor} and \eqref{eq:un_cor} in \eqref{ll5} and from \eqref{eq:multi_scale_last}, we get negative log-likelihood for pair $(x_a, x_b)$ for the scale $L$ as
\begin{align}
    \ell_{F}(\mathbf{h}_{L-1_a},\mathbf{h}_{L-1_b};F) = &\sum_{k=1}^{K}||\mathbf{z}_{L_a}^k||^2 + \log \left|f'_L(\mathbf{h}_{L-1_a})\right| + \sum_{k\neq F} ||\mathbf{z}_{L_b}^k||^2\nonumber\\
    &+ \log \left|f'_L(\mathbf{h}_{L-1_b})\right| + \frac{\left|\left|\mathbf{z}_{L_b}^F - \sigma_{ab}\mathbf{z}_{L_a}^F\right|\right|^2}{1-\sigma_{ab}^2}\label{eq:l_f}
\end{align}

The loss $\ell_F$, also called the Factor Loss is responsible for obtaining disentangled representations in the final latent $\mathbf{z}_L$. $\sigma_{ab} \in (0,1)$ in Equation \ref{eq:l_f} is called the positive correlation coefficient. The closer $\sigma_{ab}$ is to $1$, the stronger the factorization is.
The scales $1\dotsb L-1$ are optimized for each image independently. From \eqref{eq:flow_loglikelihood}, \eqref{eq:multi_scale_rep}, and \eqref{eq:z1l_unit} the negative log-likelihood for each scale $i \in \{1,2,\dotsb,L-1\}$,
\begin{align}
    \ell_i(\mathbf{h}_{i-1}) &= 0.5\left(\log(2\pi) + ||f_i(\mathbf{h}_{i-1})||^2\right)+\log\left|\det f'_i(\mathbf{h}_{i-1})\right|\nonumber\\
    &= 0.5\left(\log(2\pi) + ||\mathbf{z}_i||^2\right)+ \log \left|\det f'_i(\mathbf{h}_{i-1})\right|\label{eq:l_i}
\end{align}

and the loss per sample for the intermediate scales,
\begin{align}
    \ell_s(\mathbf{h}_0) &= \sum_{i=1}^{L-1} 0.5\left(\log(2\pi) + ||\mathbf{z}_i||^2\right) + \log \left|\det f'_i(\mathbf{h}_{i-1})\right|\label{eq:l_s}
\end{align}

Combining \eqref{eq:l_f} and \eqref{eq:l_s}, the total loss for a given pair of images $(\vec{x}_a,\vec{x}_b)\sim p(\vec{x}_a,\vec{x}_b;F)$ is
\begin{align}
    \ell(\mathbf{h}_{0_a}, \mathbf{h}_{0_b}; F) &= \ell_s(\mathbf{h}_{0_a}) + \ell_s(\mathbf{h}_{0_b}) +\ell_F(\mathbf{h}_{L-1_a},\mathbf{h}_{L-1_b};F) \label{eq:total_loss}
\end{align}
where $\mathbf{h}_{0_a} = \mathbf{x}_a$ and $\mathbf{h}_{0_b}=\mathbf{x}_b$. 
Figure \ref{fig:glowin_loss} shows a pictorial representation of the loss functions. As can be seen, the Factor Loss $\ell_F$ responsible for the disentanglement is applied between the final latents of images $\mathbf{x}_a$ and $\mathbf{x}_b$, while the scale loss $\ell_s$ which is just  the negative log-likelihood is optimized for the intermediate latents for both the input images separately.

\section{Experiments and Results}

\subsection{Dataset}
We conduct experiments on the BraTS 2019 dataset \cite{menze2014multimodal,bakas2017advancing, bakas2019identifying} with 335 co-registered, skull-stripped Brain MRI scans of pre-operative glioblastoma and lower grade glioma with ground truth segmentations.
The dataset is split into training (29976 slices, 217 studies), validation (7038 slices, 51 studies) and test (9868 slices, 67 studies) sets.  
In the experiments, we use slices from the axial plane in the FLAIR modality. We use presence of anomalies(from ground truth segmentation) and the slice index in each slice as the generative factors to train our model. The model was written using PyTorch 1.4 and python 3.6 and trained using 4 RTX 2070 GPUs.

\subsection{Qualitative Experiments}
For the experiments, we follow a similar multi-scale architecture to Glow \cite{kingma2018glow}, with $L=3$ and $K=16$, use $1\times1$ invertible convolutions and affine coupling. We train our model on $64\times64$, single channel images. Table \ref{tab:factor_dims} shows the dimensions for the factors we used for training the model. 
We compare our model with the baseline Glow \cite{kingma2018glow} and VAE based IIN \cite{esser2020disentangling} models.

\begin{table}[t]
\caption{Factor dimensions used for training the GLOWin and IIN models.}\label{tab:factor_dims}
\centering
\begin{tabular}{|l|l|l|}
\hline
\bfseries Factor & \bfseries IIN & \bfseries GLOWin\\
\hline
  Anomaly ($\vec{z}_L^{ano}$) & $384$ & $6\times 8\times 8 \;(384)$\\
  Slice Index ($\vec{z}_L^{slx}$)& $512$ & $8\times 8\times 8 \;(512)$\\
  Residual ($\vec{z}_L^{res}$)& $128$ & $2\times 8\times 8 \;(128)$\\
\hline
\end{tabular}
\end{table}

\begin{figure}[t]
\centering
\includegraphics[width=0.7\linewidth]{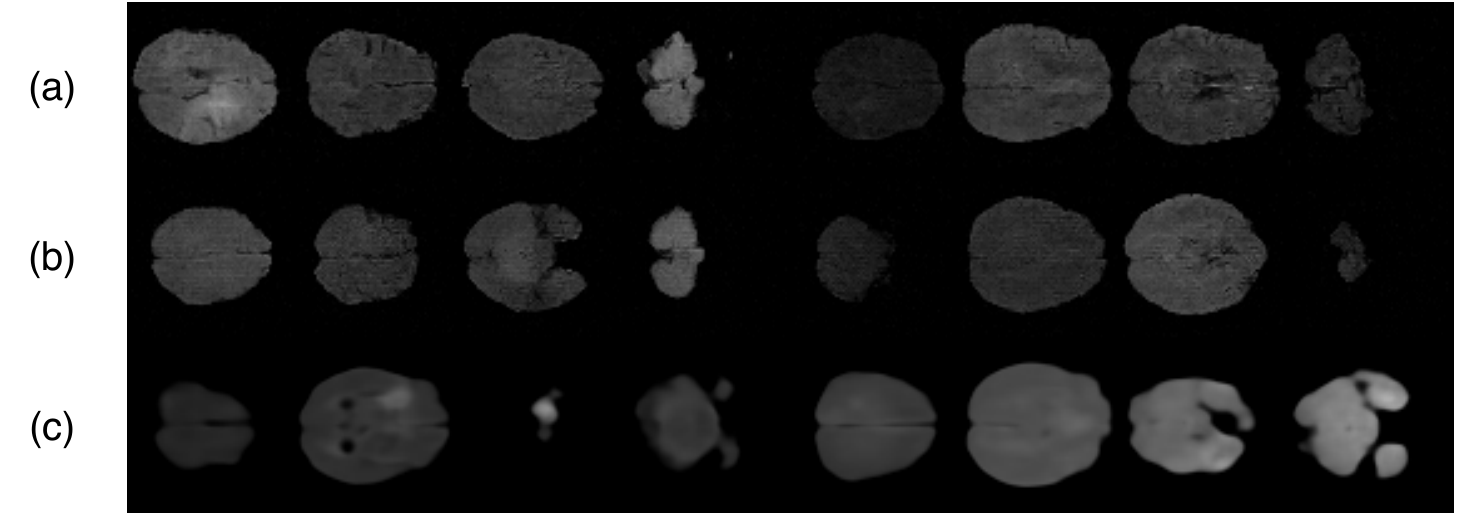}
\caption{Random images generated by the GLOWin (a), Glow (b) and IIN (c) models with samples drawn at a temperature $T=0.7$.} \label{fig:rand_gen}
\end{figure}

\subsubsection{Image synthesis.} Figure \ref{fig:rand_gen} shows some random samples generated by our model, the Glow model \cite{kingma2018glow} and the IIN model \cite{esser2020disentangling}. It can be seen that our model is able to generate a variety of  realistic images with and without pathologies by drawing random samples from $\mathcal{N}(\mathbf{0},\mathbf{I})$. The Glow and GLOWin models generate more realistic samples than the IIN model. This is expected as the generative capability of the IIN model is limited by its VAE resulting in blurrier images. The samples for all the models were drawn at a temperature of $0.7$ as defined in \cite{kingma2018glow}. 

\subsubsection{Semantically meaningful interpolations.} In addition to this, because our model factorizes $z_L$ into disentangled factors, we are also able to interpolate selected factors $z_L^i$ between the images to manipulate the features represented by that factor. 
\begin{figure}[t]
\centering
\includegraphics[width=0.9\linewidth]{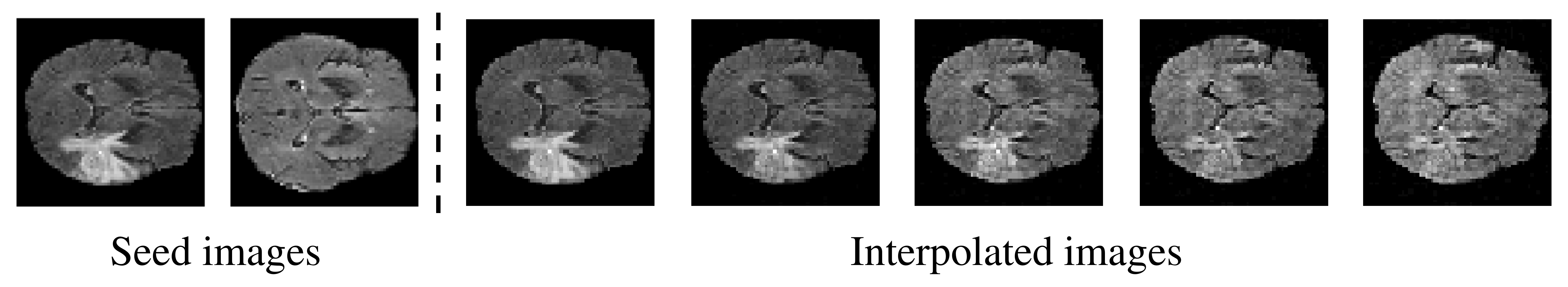}
\caption{Interpolation of $\mathbf{z}_L^{ano}$ factor.} \label{fig:fac_ano_interp}
\end{figure}
\begin{figure}[t]
\centering
\includegraphics[width=0.9\linewidth]{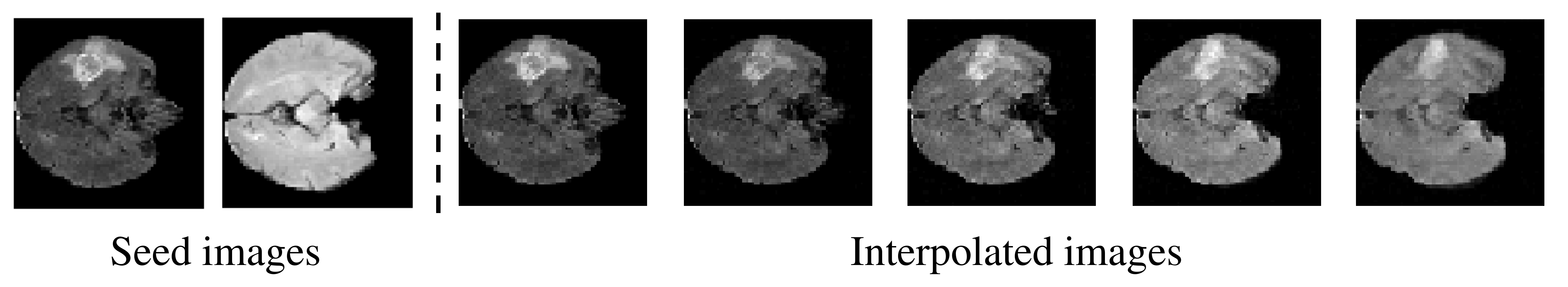}
\caption{Interpolation of $\mathbf{z}_L^{slx}$ factor.} \label{fig:fac_slx_interp}
\end{figure}
Figure \ref{fig:fac_ano_interp} shows the interpolation of the anomaly factor $z_L^{ano}$ between the two seed images. It can be seen that the interpolation of $z_L^{ano}$ from the first seed image to the second seed image removes the anomaly while maintaining the approximate shape features. Only $z_L^{ano}$ is interpolated, other factors and latents are kept fixed as the latents and factors of the first seed image.
Figure \ref{fig:fac_slx_interp} shows the interpolation of the slice index factor $z_L^{slx}$ between the two seed images. It can be seen that the interpolation of $z_L^{slx}$ from the first seed image to the second seed image changes the shape while maintaining the anomaly. Only $z_L^{slx}$ is interpolated, other factors and latents are kept fixed as the latents and factors of the first seed image. This shows that our model is able to disentangle the factors which can be modified independent of each other.

\begin{table}[t]
\centering
\makebox[0pt][c]{\parbox{1\textwidth}{%
    \begin{minipage}[b]{0.48\hsize}\centering
    \caption{Comparision of the anomaly classification downstream task.}\label{tab:ds_ano}
    \begin{tabular}{|l|l|l|l|}
    \hline
    \bfseries Model & \bfseries $z$ dim & \bfseries Accuracy & \bfseries AUC\\
    \hline
     IIN & 384 & 70.62 & 0.78\\
     Glow & 1024 & 77.04 & 0.85\\
     \textbf{GLOWin}\textbf{(ours)}& 384 & \textbf{77.23} & \textbf{0.85}\\
    \hline
    \end{tabular}
    \end{minipage}
    \hfill
    \begin{minipage}[b]{0.48\hsize}\centering
    \caption{Comparision of the slice index regression downstream task.}\label{tab:ds_reg}
    \begin{tabular}{|l|l|l|l|}
    \hline
      \bfseries Model & \bfseries $z$ dim & \bfseries MAE & \bfseries $R^2$\\
    \hline
    IIN & 512 & 0.066 & 0.847\\
    Glow & 1024 & 0.065 & 0.816\\
    \textbf{GLOWin}\textbf{(ours)}& 512 & \textbf{0.048} & \textbf{0.868}\\
    \hline
    \end{tabular}
    \end{minipage}
    }}
\end{table}

\subsection{Quantitative Experiments}
Next, we move on to the quantitative experiments where we quantitatively compare the performance of our model with the baseline models. We aim to understand the effect enforcing disentanglement has on our model. 
We compare the bits/dimension\cite{van2014student, theis2015note} achieved by our model and the Glow model \cite{kingma2018glow}. The bits/dimension are \textbf{2.588} in Glow model and \textbf{2.602} in our model(GLOWin). Even though the Glow model achieves a slightly better score, it can be seen that it is only by a very small margin. This implies that enforcing disentanglement does not affect the generative capability of the model much.
\subsubsection{Downstream tasks.} To evaluate the quality of the latent space, we perform two downstream tasks: anomaly classification and slice index regression. 
The anomaly classification task is designed to verify how the model encodes the latent feature regarding the presence of pathologies. The slice index regression task is designed to verify how well the model learns the anatomical location and shape of brain images. We compare the results between our model Glow and IIN in terms of classification accuracy and area under the receiver operating characteristic curve (AUC) for the classification task and Mean Absolute Error (MAE) and coefficient of determination ($R^2$) for the regression task. For the Glow and GLOWin models, we only use the flattened final latent $z_L$ for training the classifier and regression models. Further to compare degree of the disentanglement with IIN, we only use $z_{L_{GLOWin}}^{ano}$ and $z_{IIN}^{ano}$ for training the classification models and $z_{L_{GLOWin}}^{slx}$ and $z_{IIN}^{slx}$ for training the slice index regression models. Random Forest Classifier was used for the classification task and a 4 layer MLP was used for the regression task. As seen in Table \ref{tab:ds_ano}, GLOWin achieves a similar accuracy and AUC as Glow while using smaller representations with both models having higher scores than IIN. Table \ref{tab:ds_reg} hints that GLOWin may outperform both Glow and IIN in the regression task while again using less features than Glow.

\begin{figure}[htb]
\centering
\includegraphics[width=0.6\linewidth]{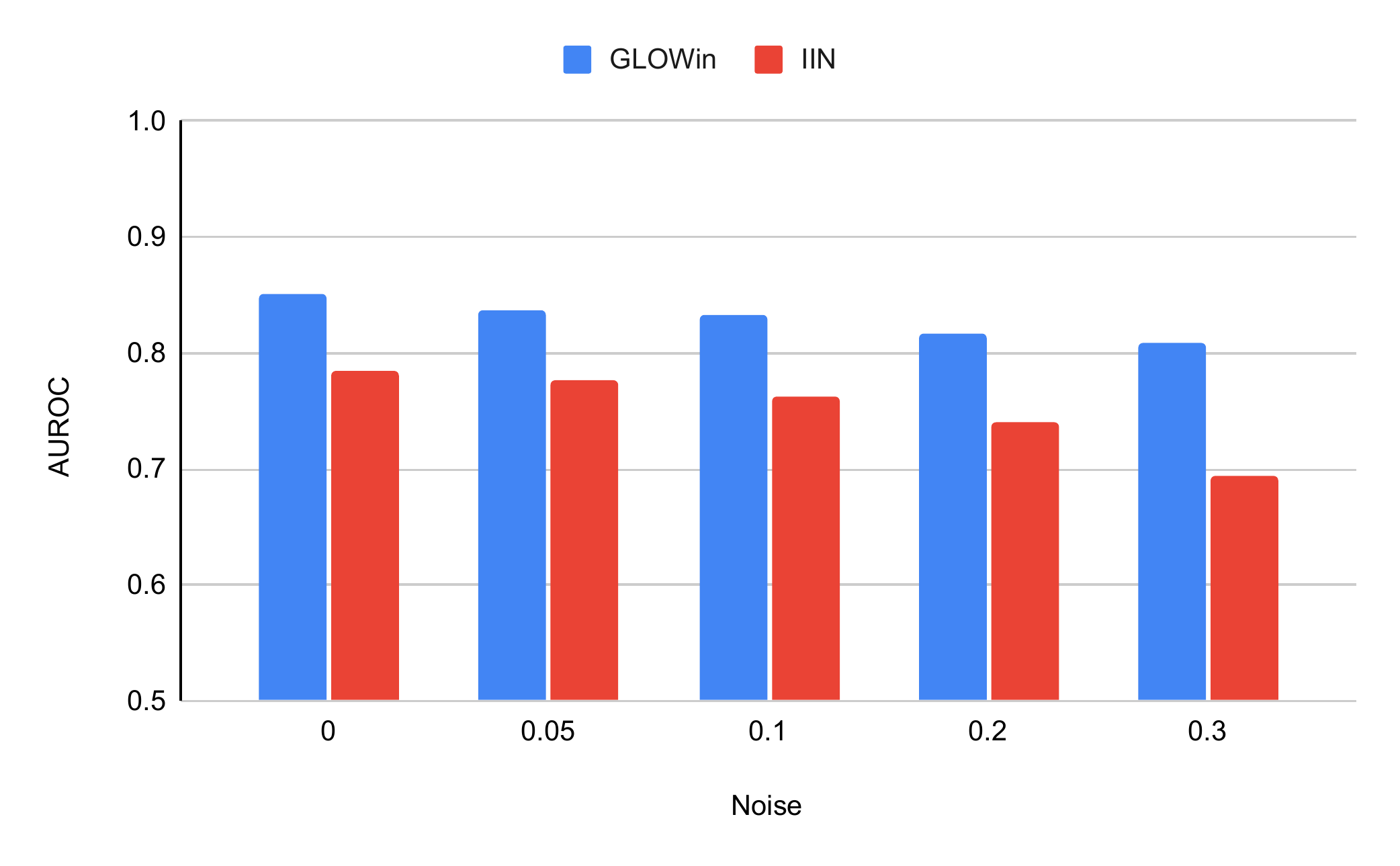}
\caption{Accuracy of downstream anomaly classification according to the different noise levels.} \label{fig:noise_auc}
\end{figure}

\subsubsection{Robustness to Noise} To evaluate the robustness of the representations to noise, we evaluated the impact of adding noise to input images during test time of the anomaly classification task. For this, random unit gaussian noise weighted at $[0, 0.05, 0.1, 0.2, 0.3]$ was added to the input image. From Figure \ref{fig:noise_auc}, we can see that while the AUC for our GLOWin model decreases, it is not as affected by noise as the baseline IIN model. 

\section{Discussion and Conclusion}
In this work we have proposed a flow-based generative framework called GLOWin and showcased its ability to learn disentangled representations in the latent space. In the quantitative and qualitative experiments, we have seen that our model is not only able to disentangle semantic concepts into factors in the latent space, but is also able to achieve comparable or better scores than the baseline models in the downstream tasks. The end-to-end invertible nature of GLOWin, allows utilizing these factors to generate realistic images not suffering from the blurriness of its VAE counterpart.
To the best of our knowledge, this is the first time a flow-based generative model has been used to learn disentangled representations in the medical imaging domain. Compared to existing works in the medical imaging domain that learn separate latent spaces for each disentangled factor using separate encoders and make use of a combination of several loss functions, our model is able to learn disentangled representations by factorizing the same latent space into independent factors. Another advantage of our model is that it just requires image pairs containing the same semantic concepts for training. It also does not require explicit labels.
Our framework also comes with its set of limitations. First, the flow-based models are unable to perform dimensionality reduction like the VAEs, so the latent dimensions are the same dimensions as the input dimension. This restricts the size of the image we can train our models on. The design choice to enforce disentanglement in the embeddings from the last scale of the multi-scale architecture also means that features represented by the intermediate latents are not considered for disentanglement and thus could affect the disentanglement ability of our model. In the future, we seek to conduct more comprehensive experiments with more datasets and factors to evaluate the representations learnt by our model further and explore its application for data augmentation of underrepresented combinations of factors as well as domain adpatation and interpretability.

\newpage
\appendix
\section{Appendix}
\subsection{Reconstruction of Noisy Images}

\begin{figure}[htb]
\centering
\includegraphics[width=0.8\linewidth]{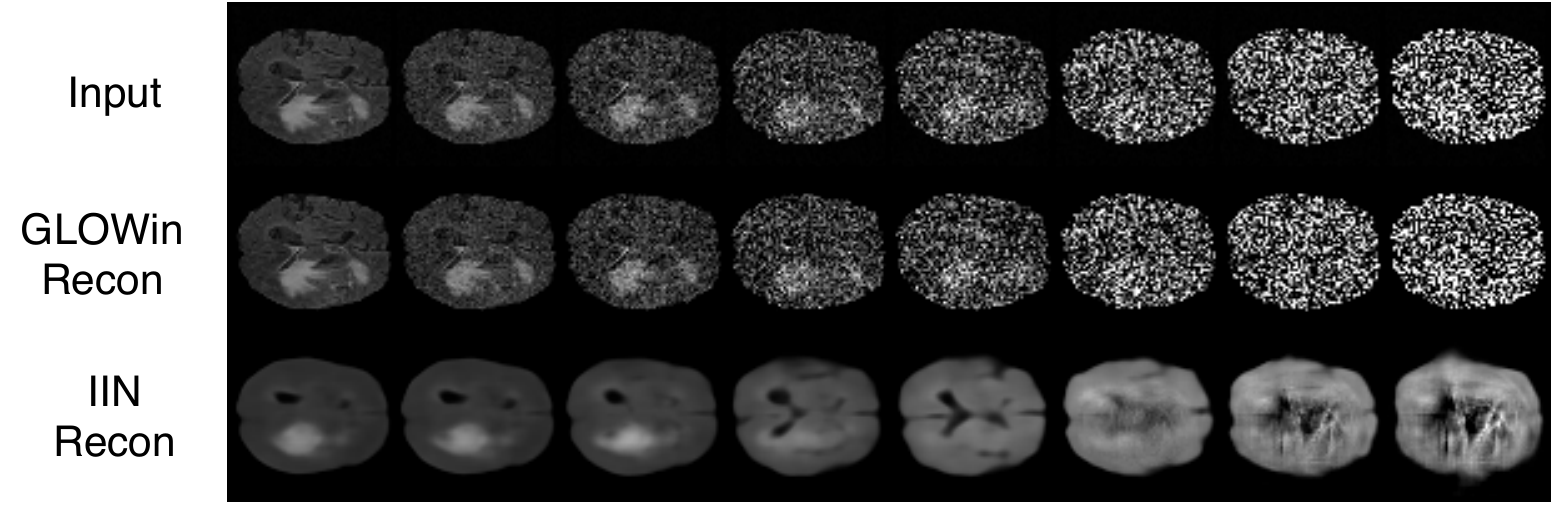}
\caption{Reconstruction of images with random noise weighted at 0, 0.05, 0.1, 0.2, 0.3, 0.5, 0.7 \text{and} 1.0. First Row: Input, Second Row: GLOWin Reconstruction, Third Row: IIN Reconstruction.} \label{fig:recon}
\end{figure}

\subsection{Interpolation between two samples}
\begin{figure}[htb]
\centering
\includegraphics[width=0.5\linewidth]{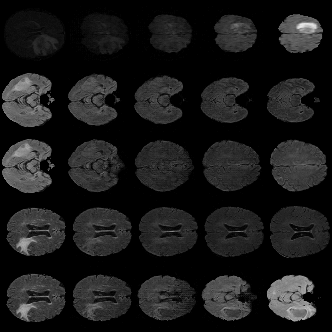}
\caption{Interpolation between two real samples. The first and last columns are real samples, the middle columns are the interpolated images.} \label{fig:recon}
\end{figure}
\newpage

\subsection{Architecture Details}

\begin{table}[!htb]
\centering
\makebox[0pt][c]{\parbox{1\textwidth}{%
    \begin{minipage}[b]{0.48\hsize}\centering
    \caption{Parameters used for training Glow and GLOWin.}\label{tab:ds_ano}
     \begin{tabular}{|l|l|}
      \hline
      \bfseries Parameter & \bfseries Value\\
      \hline
      L & 3\\
      K & 16\\
      M & 2 \\
      n\textunderscore bits & 6\\
      Batch Size & 80\\
      Input Dimension & $1\times 64\times 64$\\
      $\sigma_{ab}$(GLOWin) & 0.85\\
      Learning Rate & $1e-4$\\
      Temperature & 0.7\\
      Optimizer & Adam\\
      Epochs & 100\\
      \hline
      \end{tabular}
    \end{minipage}
    \hfill
    \begin{minipage}[b]{0.48\hsize}\centering
      \caption{Parameters used for training IIN.}
      \centering
      \begin{tabular}{|l|l|}
      \hline
      \bfseries Parameter & \bfseries Value\\
      \hline
      n$_{flow}$ & 12\\
      D & 2\\
      H & 512\\
      Batch Size & 256\\
      Input Dimension & $1\times 64\times 64$\\
      $\sigma_{ab}$ & 0.85\\
      Learning Rate & $1e-5$\\
      Temperature & 0.7\\
      Optimizer & Adam\\
      Epochs & 60\\
      \hline
      \end{tabular}
    \end{minipage}
    }}
\end{table}

\subsection{GLOWin: Training Algorithm}
\begin{algorithm}[H]
\SetAlgoLined
\SetKwProg{Init}{Initialize}{}{}
\KwData{ $\mathbf{X}_a, \mathbf{X}_b, \mathrm{Factor}\,\mathbf{F}$}
\textbf{Initialize:} parameters for the GLOWin model $g_{\theta}$.\\
\For{$\mathbf{x}_a, \mathbf{x}_b, F$ in \textbf{Data}}{
    $\mathbf{z}_a \leftarrow g_{\theta}(\mathbf{x}_a + UniformNoise(\frac{1}{2^{n\_bits}}))$\;
    $\mathbf{z}_b \leftarrow g_{\theta}(\mathbf{x}_b + UniformNoise(\frac{1}{2^{n\_bits}}))$\;
    $loss_{s_a} \leftarrow \ell_s(\mathbf{z}_{{1 \dots L-1}_a})$\;
    $loss_{s_b} \leftarrow \ell_s(\mathbf{z}_{{1 \dots L-1}_b})$\;
    $loss_{F_{ab}} \leftarrow \ell_F(\mathbf{z}_{L_a}, \mathbf{z}_{L_b}, F)$\;
    $loss_{ab} \leftarrow loss_{s_a} + loss_{s_b} + loss_{F_{ab}}$\;
    \textbf{Optimize: }$loss_{ab}$ using Adam optimizer
 }

\caption{Algorithm for training the GLOWin model.}
\label{alg:glowin}
\end{algorithm}

\bibliographystyle{splncs04}
\bibliography{bibliography}

\begin{thebibliography}{10}
\providecommand{\url}[1]{\texttt{#1}}
\providecommand{\urlprefix}{URL }
\providecommand{\doi}[1]{https://doi.org/#1}

\bibitem{bakas2017advancing}
Bakas, S., Akbari, H., Sotiras, A., Bilello, M., Rozycki, M., Kirby, J.S.,
  Freymann, J.B., Farahani, K., Davatzikos, C.: Advancing the cancer genome
  atlas glioma mri collections with expert segmentation labels and radiomic
  features. Scientific data  \textbf{4},  170117 (2017)

\bibitem{bakas2019identifying}
Bakas, S., Reyes, M., Jakab, A., Bauer, S., Rempfler, M., Crimi, A., Shinohara,
  R.T., Berger, C., Ha, S.M., Rozycki, M., et~al.: Identifying the best machine
  learning algorithms for brain tumor segmentation, progression assessment, and
  overall survival prediction in the brats challenge. arXiv preprint
  arXiv:1811.02629  (2018)

\bibitem{bengio2013representation}
Bengio, Y., Courville, A., Vincent, P.: Representation learning: A review and
  new perspectives. IEEE transactions on pattern analysis and machine
  intelligence  \textbf{35}(8),  1798--1828 (2013)

\bibitem{Chartsias_2019}
Chartsias, A., Joyce, T., Papanastasiou, G., Semple, S., Williams, M., Newby,
  D.E., Dharmakumar, R., Tsaftaris, S.A.: Disentangled representation learning
  in cardiac image analysis. Medical image analysis  \textbf{58},  101535
  (2019)

\bibitem{chen2018isolating}
Chen, R.T., Li, X., Grosse, R.B., Duvenaud, D.K.: Isolating sources of
  disentanglement in variational autoencoders. Advances in Neural Information
  Processing Systems  \textbf{31},  2610--2620 (2018)

\bibitem{dinh2014nice}
Dinh, L., Krueger, D., Bengio, Y.: Nice: Non-linear independent components
  estimation. ICLR  (2015)

\bibitem{dinh2016density}
Dinh, L., Sohl-Dickstein, J., Bengio, S.: Density estimation using real nvp.
  ICLR  (2017)

\bibitem{esser2020disentangling}
Esser, P., Rombach, R., Ommer, B.: A disentangling invertible interpretation
  network for explaining latent representations. CVPR pp. 9223--9232 (2020)

\bibitem{goodfellow2014generative}
Goodfellow, I., Pouget-Abadie, J., Mirza, M., Xu, B., Warde-Farley, D., Ozair,
  S., Courville, A., Bengio, Y.: Generative adversarial nets. Advances in
  Neural Information Processing Systems pp. 2672--2680 (2014)

\bibitem{betavae}
Higgins, I., Matthey, L., Pal, A., Burgess, C., Glorot, X., Botvinick, M.,
  Mohamed, S., Lerchner, A.: beta-vae: Learning basic visual concepts with a
  constrained variational framework. ICLR  (2017)

\bibitem{jacobsen2018revnet}
Jacobsen, J.H., Smeulders, A., Oyallon, E.: i-revnet: Deep invertible networks.
  arXiv preprint arXiv:1802.07088  (2018)

\bibitem{jiang2020semisupervised}
Jiang, H., Chartsias, A., Zhang, X., Papanastasiou, G., Semple, S., Dweck, M.,
  Semple, D., Dharmakumar, R., Tsaftaris, S.A.: Semi-supervised pathology
  segmentation with disentangled representations. Domain Adaptation and
  Representation Transfer, and Distributed and Collaborative Learning pp.
  62--72 (2020)

\bibitem{kim2018disentangling}
Kim, H., Mnih, A.: Disentangling by factorising. ICML  (2018)

\bibitem{kingma2013autoencoding}
Kingma, D.P., Welling, M.: Auto-encoding variational bayes. arXiv preprint
  arXiv:1312.6114  (2013)

\bibitem{kingma2018glow}
Kingma, D.P., Dhariwal, P.: Glow: Generative flow with invertible 1x1
  convolutions. Advances in Neural Information Processing Systems  (2018)

\bibitem{Litjens_2017}
Litjens, G., Kooi, T., Bejnordi, B.E., Setio, A.A.A., Ciompi, F., Ghafoorian,
  M., Van Der~Laak, J.A., Van~Ginneken, B., S{\'a}nchez, C.I.: A survey on deep
  learning in medical image analysis. Medical image analysis  \textbf{42},
  60--88 (2017)

\bibitem{menze2014multimodal}
Menze, B.H., Jakab, A., Bauer, S., Kalpathy-Cramer, J., Farahani, K., Kirby,
  J., Burren, Y., Porz, N., Slotboom, J., Wiest, R., et~al.: The multimodal
  brain tumor image segmentation benchmark (brats). IEEE transactions on
  medical imaging  \textbf{34}(10),  1993--2024 (2014)

\bibitem{van2014student}
van~den Oord, A., Schrauwen, B.: The student-t mixture as a natural image patch
  prior with application to image compression. J. Mach. Learn. Res.
  \textbf{15}(1),  2061--2086 (2014)

\bibitem{qin2019unsupervised}
Qin, C., Shi, B., Liao, R., Mansi, T., Rueckert, D., Kamen, A.: Unsupervised
  deformable registration for multi-modal images via disentangled
  representations. International Conference on Information Processing in
  Medical Imaging  (2019)

\bibitem{theis2015note}
Theis, L., Oord, A.v.d., Bethge, M.: A note on the evaluation of generative
  models. arXiv preprint arXiv:1511.01844  (2015)

\end{thebibliography}

\end{document}